\documentclass[letterpaper]{article} 
\usepackage{aaai2026}  
\usepackage{times}  
\usepackage{helvet}  
\usepackage{courier}  
\usepackage[hyphens]{url}  
\usepackage{graphicx} 
\usepackage{natbib}  
\usepackage{caption} 
\usepackage{algorithm}
\usepackage{algorithmic}

\usepackage{newfloat}
\usepackage{listings}
\DeclareCaptionStyle{ruled}{labelfont=normalfont,labelsep=colon,strut=off} 
\floatstyle{ruled}
\newfloat{listing}{tb}{lst}{}
\floatname{listing}{Listing}
\usepackage{graphicx}
\usepackage{subcaption}

\title{See Me, Believe Me: Causality, Intersectionality, and Interventions Improving the Appearance of Patients}
\author {
    Kenya S. Andrews\textsuperscript{\rm 1},
    Mesrob I. Ohannessian\textsuperscript{\rm 2},
    Elena Zheleva\textsuperscript{\rm 2}
}
\affiliations {
    \textsuperscript{\rm 1}Brown University,
    \textsuperscript{\rm 2}University of Illinois Chicago\\
    kenya\_andrews@brown.edu, mesrob@uic.edu, ezheleva@uic.edu 
}

\usepackage{bibentry}

\begin{document}

\maketitle

\begin{abstract}
In the context of medical records, patients often experience testimonial injustice, where the textual account undermines the validity of their experiences. Past work has demonstrated that intersectionality of demographic features is crucial to \emph{detect} such injustice. We use causal discovery to study the degree to which certain demographic features tied to marginalization (namely age, gender, and race), together lead to specific types of testimonial injustice terms. This offers us an insightful Structural Causal Model (SCM) relating those demographic features to the different ways a patients' reality may be undermined. We then move toward \emph{addressing} such injustice, by very selectively (based on insights from the SCM) editing physicians' notes. For comparison, we contrast these rule-based edits with blanket context-based modifications using an LLM. We assess the impact of these changes on the perception of patients' experiences, using human experts and an LLM. We find that edits, in general, enhance clarity regarding the urgency and causes of patients' conditions. Additionally, for human experts, rule-based modifications show a tendency to shift blame away from patients, and toward more objective external factors. These findings (1) underscore the importance of quantifying sources of injustice in how patients' testimonies are recorded, (2) reveal that making minimal intentional changes accordingly can effect improved patient perception, and (3) call for larger efforts to assess health outcomes under such more just representation.
\end{abstract}


\section{Introduction}
\label{sec:intro}
Patients seeking medical treatment are not only vulnerable but are simultaneously dependent upon whomever is giving them care at the time. This fact is particularly concerning for those who are not believed or appropriately perceived because of prejudices about them, an experience known as testimonial injustice \cite{fricker2019testimonial}. It has been proven that clinicians are more likely to ignore and make light of the concerns of Black and female patients than White and male patients \cite{beach2021testimonial, chitribune}. 
Yet, very little work has been done to show the nuances of the experiences for younger Black females, younger Black males, senior White males, senior Latina females, and those of other intersections. \emph{Intersectionality} informs on how people experience the world due to their attributes, such as demographic features (e.g., race, age, gender) \citep{crenshaw2013mapping,marques2018patricia}. We hypothesize that a key measurable contributor to these nuanced experiences in healthcare might be testimonial injustice. 

In such critical circumstances, the trust that patients place in clinical staff to be heard and understood becomes ever more pertinent. However, this trust can be compromised by deeply rooted biases that may permeate the staff's language (i.e., word choices) in patient records (i.e., physicians' notes). The choice of words in clinical notes often reflects biases, which can skew understanding and lead to disparate patient outcomes for readers of the notes (including Machine Learning (ML) informed tools), leading to injustices. Further, as Large Language Models (LLMs) are more heavily incorporated into clinical settings, it becomes imperative that we scrutinize the linguistic choices that will inform these models and shape their perceptions of patients. Both clinicians and LLMs risk absorbing these biases, potentially influencing their clinical judgments, which can compromise accuracy and effectiveness in patient outcomes.

Though it is known that intersectionality of features is necessary to uncover certain cases of testimonial injustice \cite{andrews-etal-2023-intersectionality}, we lack a comprehensive understanding of \emph{which} features causally lead to testimonial injustice, \emph{how} they do so, and their levels of impact. This gap in knowledge prevents exploring if \emph{intentional interventions}  can lead to changes within patient experiences and outcomes.

Here, we aim to address this by using causal discovery to not only show how attributes of a person can come together --- i.e., intersectionality --- to contribute to them experiencing testimonial injustice, but also to understand the specific ways injustice is revealed, the intensity with which attributes are contributors, test strategic interventional methods along the intersectional paths of injustice, and understand the impact of realigning descriptions of conditions for patients who fall along such paths.

Thus, the research questions we explore here are: (RQ1) \textbf{Can we identify how individual demographic features influence language in medical settings, leading collectively to testimonial injustice?} And, once we do, (RQ2) \textbf{can we quantify the degree to which these interactions contribute to this experience to make intentional interventions?}

The contributions of this paper are: (1) using causal discovery methods and the resulting Structural Causal Model (SCM) to identify and describe how demographic features of concern (i.e., race, age, and gender) contribute to a patient experiencing testimonial injustice in physicians' notes,
(2) discussing the revealed interactions and quantifying the extent of their influence on testimonial injustice, (3) investigating a use-case study for these insights through intentional intervention methods (i.e., modifications to physicians' notes) to address injustices in physicians' notes based on the likelihood of such experiences as indicated by the SCM and context-specific adjustments, and (4) assessing such interventions by comparing the downstream effects of unmodified vs. modified notes, specifically in how patients and their conditions are perceived both by human experts and LLMs.

\section{Related Works}

An empirical study conducted by \citet{andrews-etal-2023-intersectionality} showed that there is differential treatment between subgroups experiencing testimonial injustice in medical settings, noting that these nuances can only be revealed through the lens of intersectionality. However, the causal nature of this intersectionality was left open, namely, regarding which attributes contribute to testimonial injustice and the degree to which these attributes influence someone experiencing such injustice.
Thus, this work shifts the focus to how one's demographic features--- namely race, gender, and age ---can contribute to them experiencing testimonial injustice, using causal discovery. To the best of our knowledge, ours is the first use of causal discovery to understand these nuanced experiences of testimonial injustice in medical settings through an intersectional lens. 

\citet{amemiya2023thinking} developed a framework which tells how people attribute inequality to structural causes, namely instances in society that systematically advantage some and marginalize others. This is rampant in the medical field \citep{hall2015implicit}, with various instances of preferential treatment to those with specific insurance policies, race, income, etc. \citep{stepanikova2008effects, yearby2022structural}. They showed that when 2 groups have the same abilities, but are systematically treated differently, there is a case for using between-group comparisons to build causal models. In medical settings, abilities are similar --- however, features can vary vastly, i.e., among those who do/do not have insurance, gender, race, age, education level, etc. Though this work focuses on observably inferrable features (i.e., race, gender, and age), we acknowledge here that there are many demographic features --- both externally observable and latent --- that can also contribute to someone experiencing testimonial injustice, particularly in medical settings. Further, the degree to which the presence of testimonially unjust terms affects care can vary due to various factors like experience. In \cite{lane1983prejudice}, registered nurses (RNs) were shown to be more susceptible to stigmatizing and judging patients when such words were present, in contrast to nursing students. The presence of such words led both students and RNs to assumptions of \emph{psychological} pain and labeling the patient as \emph{difficult}. For students, they reported less desire to work with such patients. These perceptions are likely to lead to poor outcomes for patients. Inspired by these studies, the perspective we adopt here is that, as long as there are even some clinicians whose perceptions are going to negatively affect patients, it is worthwhile to intervene and prevent it from happening.

Recent studies indicate that LLMs can significantly enhance diagnostic accuracy, nearly doubling clinicians' performance in difficult cases from the U.S. Medical Licensing Examination Step 1 Sample Exam \citep{usmle2023step1, lee2023benefits}. This suggests that even in complex cases, LLMs are strong evaluators of a patient's condition. Consequently, there has been an increase in doctors using publicly available LLMs to assist with clinical decision-making \citep{Gliadkovskaya_2024}. Further, researchers have shown that even though demographic features, specifically race, are often absent when provided to LLMs, the models are still able to infer them due to the commonality of unjust word choices in records of racial minorities \citep{hadam22}. This has led to the perpetuation of injustices in clinical treatment decisions. The concern then shifts from note readability to understanding what improvements are necessary to help LLMs interpret physicians' notes despite human bias and perceive patients more accurately.

Numerous strategies have been proposed for editing text to enhance clarity and standardization. One common approach involves neutralizing text based on individual characteristics \citep{pryzant2020automatically}. However, blindly neutralizing text is not always ideal, as it can diminish personal experiences rather than improve perception \citep{boudana2016impartiality, gonen2019lipstick}. In many cases, it can diminish the nuances of an experience or erase culturally significant aspects of an experience. Therefore, this work prioritizes building a nuanced understanding of textual influence through causal modeling, and selectively and judiciously using this insight for potential interventions to improve how patients appear to both human experts and LLMs.



\section{Data}

\subsection{MIMIC-III} \label{ssec:mimic3}
We use the \textsf{MIMIC-III} dataset as referenced in the Data Availability Statement. These records contain features of interest to the experiments conducted here, including: ethnicity/race, gender, age, patient id, diagnosis, physicians' notes, etc. (e.g., Black, female, 47, 5432, Bronchitis, ``patient claims to be experiencing...'').

\subsubsection{Pre-processing}
We identify race, gender, and age as features present that can be \emph{inferred} by people, based on visible observations. The proportion of racial groups represented in the dataset are highly imbalanced (see Table \ref{tab:distribution}), which is likely due to the region the hospital is located in. The \textsf{MIMIC-III} feature `ethnicity' often included both race and region of origin (e.g., `Asian -- VIETNAMESE'). For our analysis, we focused on race rather than region, recoding entries such as `Asian -- VIETNAMESE' simply as `Asian,' following the approach of \citet{andrews-etal-2023-intersectionality}. We removed ethnicities that were listed as ``unknown/not specified'', ``multi-race ethnicity'', ``other'', ``unable to obtain'', and ``patient declined to answer'' since they cannot be clearly associated with any race.

To address the presence of multiple records for patients, we combine the patients' records based on their patient id, gender, race, and diagnosis (e.g., 2213, male, Latino, Pancreatic Cancer). We do not combine records based on age since, in a single year, many of the patients returned for multiple visits to the ICU--- many for the same condition/ diagnosis. When we run an analysis on the physicians' notes to quantify terms that are testimonially unjust for each category of linguistic term, we aggregate counts over records for each patient and divide by the number of their records, thus obtaining the \emph{(average) number of unjust terms per record}. This allows us to ensure that no patient is more heavily weighted than another, depending on the duration of their stay or the number of doctor visits. After pre-processing, we maintain 41,886 unique patients.

\begin{table}[h]
\fontsize{7pt}{8pt}\selectfont
\centering
\begin{tabular}{l|l|l|l}
\hline
\textbf{Race} & \textbf{Gender} & \textbf{Age} & \textbf{Count} \\ \hline
Asian  & Female & Senior  & 212\\ \hline
Asian  & Female & Adult & 198\\ \hline
Asian  & Female & Child & 102 \\ \hline
Asian  & Male & Senior  & 304\\ \hline
Asian  & Male & Adult & 267\\ \hline
Asian  & Male & Child & 119 \\ \hline
Black  & Female & Senior  & 945  \\ \hline
Black  & Female & Adult & 1095  \\ \hline
Black  & Female & Child & 482 \\ \hline
Black  & Male & Senior  & 776  \\ \hline
Black  & Male & Adult & 875  \\ \hline
Black  & Male & Child & 390 \\ \hline
Latino & Female & Senior  & 81 \\ \hline
Latino & Female & Adult & 56\\ \hline
Latino & Female & Child & 27  \\ \hline
Latino & Male & Senior  & 87 \\ \hline
Latino & Male & Adult & 109\\ \hline
Latino & Male & Child & 45  \\ \hline
White  & Female & Senior  & 6076 \\ \hline
White  & Female & Adult & 6452  \\ \hline
White  & Female & Child & 2871 \\ \hline
White  & Male & Senior  & 8106 \\ \hline
White  & Male & Adult & 8496 \\ \hline
White  & Male & Child & 3715\\ \hline
\end{tabular}
\vspace{12pt}
\caption{Counts of patients by race, age, and gender} 
\label{tab:distribution}
\end{table}

\subsection{Testimonial Injustice Lexicon} 
\label{ssec: terms}
To assess testimonial injustice in physicians' notes, we focus on 4 categories of unjust terms: \textbf{evidential, judgmental, negative,} and \textbf{stigmatizing}. \textbf{Evidential} terms report a claim without asserting its truth (e.g., ``complains'', ``says''), making the patient’s experience easier to dismiss \citep{beach2021testimonial}. \textbf{Judgmental} terms convey physician skepticism (e.g., ``apparently'', ``claims''). \textbf{Negative} terms, often associated with racial and ethnic disparities \citep{neg-words}, signal denial or rejection (e.g., ``combative'', ``exaggerate'') using the lexicon from \citet{neglex}. \textbf{Stigmatizing} terms label patients via stereotypes (e.g., ``user'', ``faking'') and may influence treatment, transmit bias, and alienate patients \citep{stigmatizing,link2001conceptualizing}. Our lexicon includes conditions disproportionately affecting racial minorities, such as diabetes, substance use disorder, and chronic pain. The full based lexicon is provided in the Appendix.

\section{Causal Discovery}
To describe how inferrable demographic features (i.e., race, age, gender) can collectively lead to a patient experiencing testimonial injustice (RQ1), we perform causal discovery to build a Structural Causal Model (SCM). For this, we rely on the Fast Causal Inference (\textsf{FCI}) algorithm, which we find most suited to the present context as explained below. Before conducting our analysis, however, we first outline the assumptions that contextualize it.

\paragraph {Do you see me?} \label{sec:seeme}
We binarize our exogenous variables (i.e., gender, race, and age) according to whether an individual belongs to a marginalized group, as defined by prior empirical and sociotechnical literature.

Experiments from \citet{beach2021testimonial} show that those who are women and/or those who are Black are more likely to experience testimonial injustice compared to their White or male counterparts. Other studies have demonstrated that patients who are Black or Latino are more likely to encounter testimonial injustice in medical settings \citep{howell2018reducing}. \citet{andrews-etal-2023-intersectionality} show that the experience is much more nuanced when looking at race and gender, asserting from their experiments that Black men, Black women, Latino men, Latino women, and in some instances White women patients are more likely to experience testimonial injustice. Asian patients in ICU settings have experiences closer to that of their White counterparts and even better in some cases, \citep{zhang2020trends} --- so we do not consider them in this particular context to be a part of the marginalized races. Studies have also shown that ageism is a strong contributor to lack of proper healthcare amongst senior adults \citep{ben2017ageism}, but also for children \citep{goyal2015racial}. In the United States, how they define the age at which minors can independently consent to medical care varies. Many states allow minors to consent to certain types of care, such as sexual, mental health, or substance-related treatment, starting around age 14. Some states also permit minors as young as 14 to consent to general medical care without parental involvement, although several require minors to be 16 or older for emergency care consent without an adult present. Given these variations, we chose age 15 to capture a developmental stage where many states grant increased autonomy for medical decision-making, yet patients may still receive pediatric care.

Combining the results of these works, the demographic features that we adopt are:
\begin{itemize} \small
\itemsep-0.15em
    \item \textbf{Gender:} \texttt{\footnotesize is\_marginalized\_gender = 1} if a person is female, 
    \item \textbf{Race:} \texttt{\footnotesize is\_marginalized\_race = 1} if a person is Black or Latino, and
    \item \textbf{Age:} \texttt{\footnotesize is\_marginalized\_age = 1} if a person is a child (age $\leq 15$) or a senior (age $\geq 65$).
\end{itemize}

To study the importance of intersectionality to nuanced experiences that cannot be observed otherwise, our experiments vary the granularity of these features. In addition to the \emph{fine} features above, we also introduce a single \emph{coarse} feature that collapses the above to \texttt{\footnotesize is\_marginalized = 1}, if a person is Black male child, Black female child, adult Black man, adult Black woman, senior Black man, senior Black woman, Latino male child, Latino female child, adult Latino man, adult Latino woman, senior Latino man, senior Latino woman, White female child, adult White woman, or senior White woman, as these groups are likely to experience testimonial injustice as shown in prior work.

\paragraph{Do you believe me?}
The degree to which someone can experience testimonial injustice can vary from instance to instance (e.g., education of the listener, temperament of the listener). However, we simplify the complexity of this problem by employing a single binary outcome indicating this: \textbf{Injustice:} \texttt{\footnotesize is\_testinj = 1} if any of the categories of terms is present in ``high'' numbers. This binarization reflects the presence or absence of testimonial injustice in a given instance and does not imply that all experiences are equivalent in severity.

\subsection{Methods} \label{sec:method}

\subsubsection{Lexicon Lookup} \label{ssec: lexicon-lookup} 
Since word choice reveals attitudes one may have about a subject \citep{von2008linguistic}, we analyze the word choices of physicians in their notes to quantify patient experiences with testimonial injustice. We combine the testimonially unjust terms, into a lexicon to be used for exact-matching lookup. We \emph{expand} this lexicon by adding the stem of these words and 5 synonyms associated with each unjust term in its respective category using nltk's WordNet corpus \citep{bird2009natural}. This expansion is necessary since exact-matching is limited in reach --- there can be many variations of the same word, improperly used tenses of words, or words which are similar in meaning. This helps to more accurately cover occurrences of testimonially unjust terms.

To address the potential limitations of keyword matching, we also investigated accounting for linguistic context, namely to discount the potential uses of negation or quotation marks. Specifically, we tested for negations preceding testimonially unjust terms using the following list: [``did not", ``don't", ``will not", ``won't", ``has not", ``hasn't", ``are not", ``aren't", ``have not", ``haven't", ``do not", ``does not", ``doesn't"]. We found that out of a total of 9,353,867 testimonially unjust terms in our dataset, only 35,985 (0.38\%)  have instances of negations that occur prior to them. We also found that only 5.5\% of quoted text contains these words. Thus negations prior to testimonially unjust words and their occurrence in quotations are relatively rare and unlikely to significantly affect our causal analysis.

\subsubsection{Instances of Injustice} \label{ssec: violating}

For each of the categories of unjust terms (\textbf{Evidential}, \textbf{Judgemental}, \textbf{Negative}, \textbf{Stigmatizing}) and for each patient, we calculate the average number of unjust terms per record. Instead of representing these with their absolute numerical values, we discretize them to numbers in the range 0--9, based on which tenth quantile they belong to (i.e., 0--10$^\textrm{th}$, 10$^\textrm{th}$--20$^\textrm{th}$, ..., 90$^\textrm{th}$--100$^\textrm{th}$). Such a discretization acts as a quantization/kernel and makes the causal discovery more robust.

Ideally, our stance is that a patient is experiencing fair and just testimony when there are \emph{no} instances of unjust terms found in their records. This agrees with the perspective of \citet{andrews-etal-2023-intersectionality}. Equivalently, this means testimonial injustice occurs whenever there is presence from \emph{any} of the unjust term categories. However, to generously account for variability in context behind word choice, we loosen the notion of presence to presence with a ``high'' number of occurrences. 

We choose the 90th percentile of unjust terms per record for any single patient as such a ``high'', which, based on the above, corresponds to a numerical value of 9. Thus, we can more precisely write: \textbf{Injustice:} \texttt{\footnotesize is\_testinj = 1} if \textbf{Evidential}=9 or \textbf{Judgemental}=9 or \textbf{Negative}=9 or \textbf{Stigmatizing}=9.

\subsection{Experiments and Results} \label{sec:causresults}
\begin{figure}[h]
    \centering
    \includegraphics[height=2.25
    in]{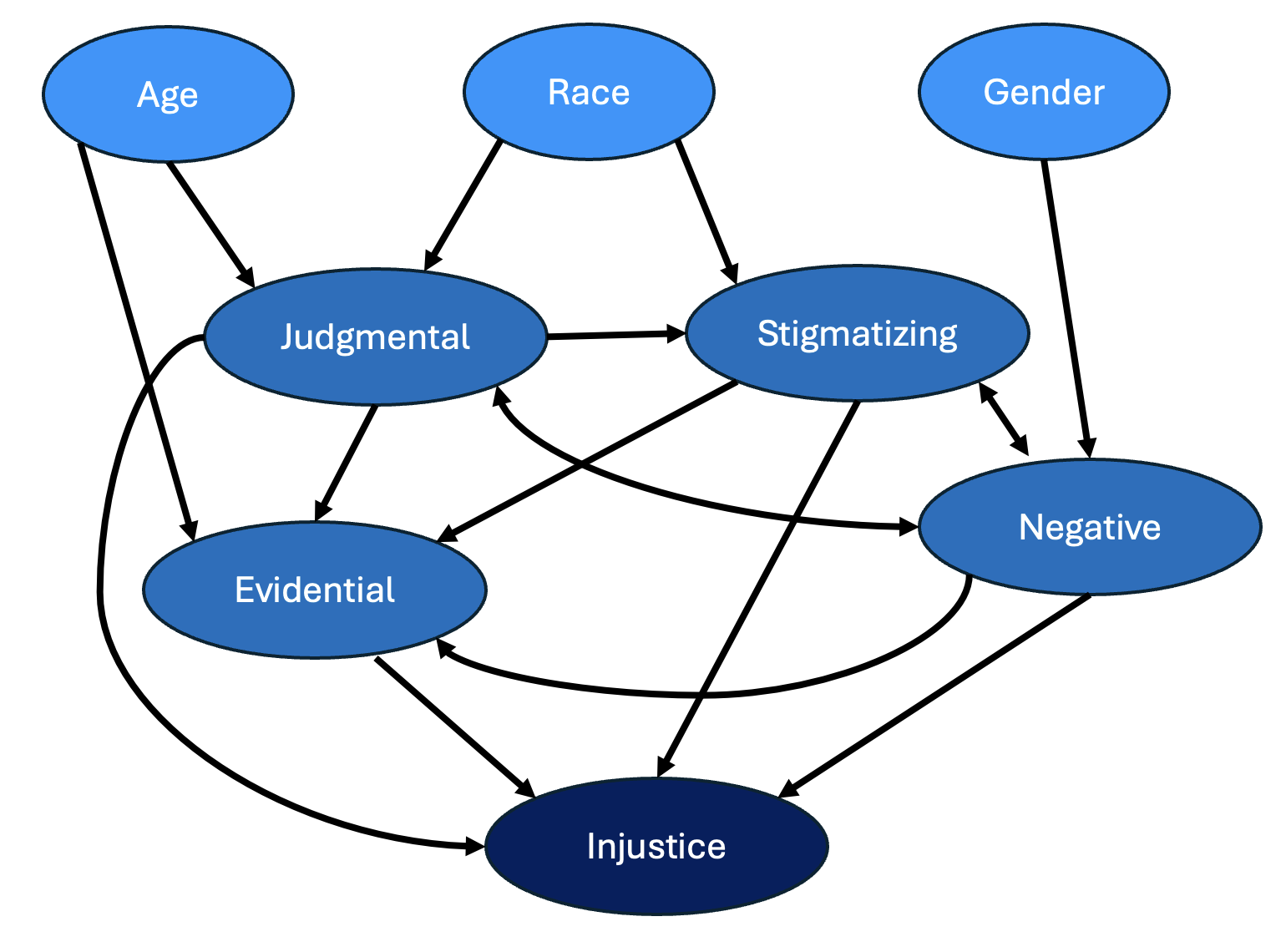}
    \caption{FCI SCMs with $\alpha = 0.05$} \label{fig:fci1}
\end{figure}

We conducted experiments to inform us of how demographic features may contribute to someone experiencing testimonial injustice, using causal discovery with the FCI algorithm. We constrain Age, Gender, and Race as exogenous variables, i.e., root causes in the graph, as we know that none of the other variables in the graph cause them. We also constrain Injustice as a leaf node, i.e., it must be an outcome of the SCM, because it can only be caused by the presencve of unjust terms. We set $\alpha = 0.05$ as the significance level threshold for the conditional independence tests used by the FCI algorithm, see Fig. \ref{fig:fci1}. Our $\alpha$ functions as a decision cutoff controlling edge removal during skeleton discovery and orientation steps. We conducted additional runs at lower and higher $\alpha$-values: $\{0.01, 0.1\}$, see Fig. \ref{fig:fci_appendix} in the Appendix.

The resulting SCM shows a direct causal edge from both race and age to judgmental terms, indicating that these factors jointly influence the outcome in an intersectional manner. The graph also reveals more subtle instances of intersectionality, mediated through the categories of testimonial injustice terms. Experiencing stigmatizing terms is affected both directly by race and indirectly by age via the mediator of judgmental terms. Similarly, experiencing evidential terms is affected both directly by age and indirectly by gender via negative terms, as well as by race via stigmatizing terms. Gender is the only feature that does not influence each and every testimonially unjust term, with judgmental terms being the only variable not connected to it. Despite the under-representation of marginalized racial and age groups in this dataset (see Table~\ref{tab:distribution}), there is a clear indication of injustice along these demographic features across all identified pathways of injustice. Even more so, since all of the features are consistently 1-hop away from experiencing testimonial injustice, we can rationalize that they are important variables in detecting if someone is likely to experience injustice. However, due to having overlapping mediators, we hypothesize that race and age particularly need to work in tandem to achieve this effect, especially when giving rise to judgmental terms.

We can also see that there exists an association between stigmatizing and negative terms that reflects a shared generative mechanism in these clinical notes, without implying a direct causal relationship between them. The same applies between judgmental and negative terms.

\begin{figure}[h]
    \centering
    \includegraphics[height=2
    in]{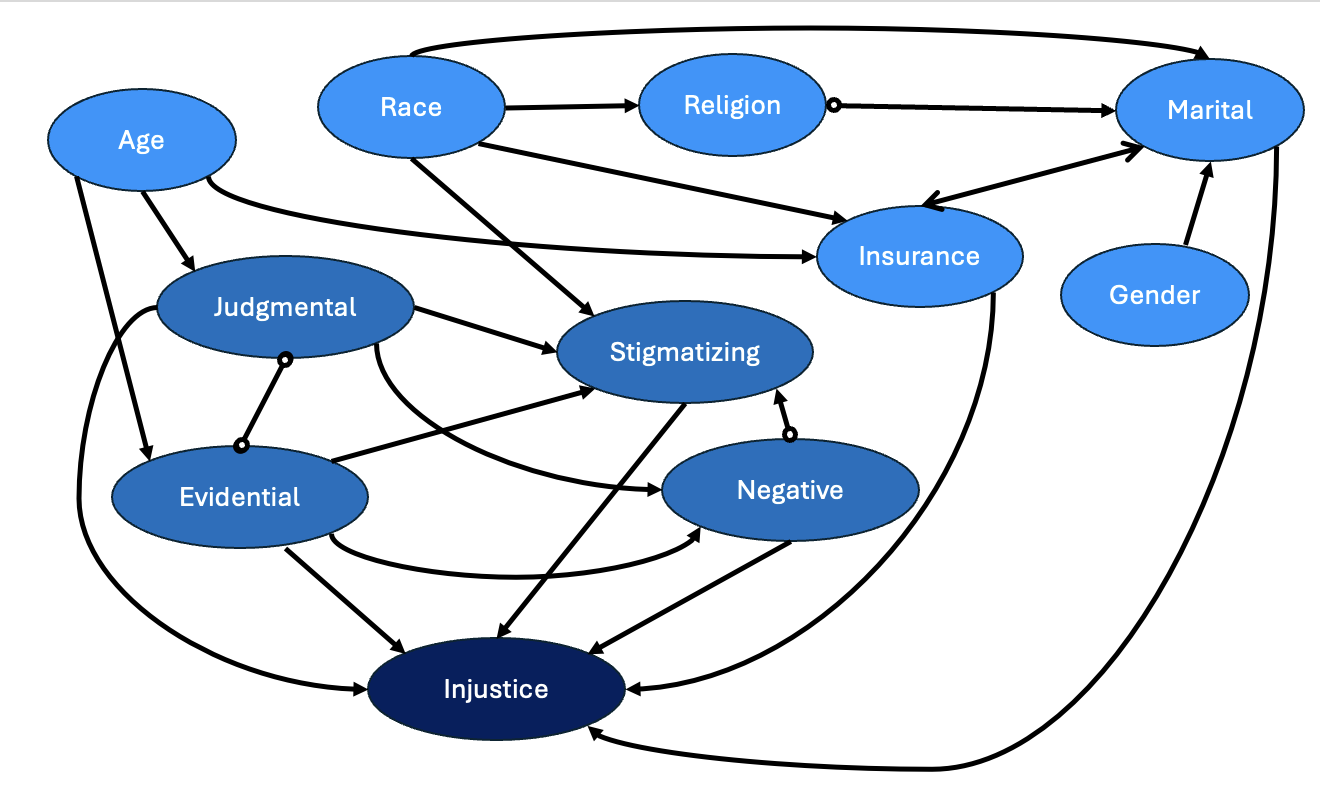}
    \caption{FCI SCMs with additional features in MIMIC-III that could act as mediators $\alpha = 0.15$} \label{fig:fci2}
\end{figure}

Though we focus on the presence of inferrable, observable demographic features, one may wonder how other features, such as insurance type (which can act as a proxy for socioeconomic status even in ICU settings \cite{monuteaux2024evaluation, cantiello2015impact}), might affect the structure of our SCM. To assess potential confounders that could significantly alter our SCM, we included all relevant factors available in MIMIC III, namely marital status, religion, and insurance type. We marked patients as likely to experience marginalization by marital status if they were single, divorced, widowed, or separated (i.e., \texttt{\footnotesize is\_marginalized\_martial = 1}), given evidence that married patients often experience lower hospital mortality, shorter length of stay, and reduced readmission rates \cite{metersky2012effect}; those likely to be marginalized by insurance type as those with non-private insurance (i.e., Medicare, Medicaid, other government, or self-pay) (i.e., \texttt{\footnotesize is\_marginalized\_insur = 1}), reflecting structural differences in access to care, as patients with government insurance may experience longer wait times and uninsured patients may have to leave without being seen \cite{soc_insurance, tolbert2024keyfacts}; and likely to experience marginalization through religious affiliation if they identify as Muslim, Jehovah’s Witness, or Hebrew/Arabic background) (i.e., \texttt{\footnotesize is\_marginalized\_religion = 1}), based on prior studies showing that Muslim patients frequently report that providers do not listen to them or question their competence, face challenges related to Islamic dress, and that providers may be reluctant to care for Jehovah’s Witnesses \cite{eriksson2023perceived, gouezec2016physicians, keshet2019language}. We note here, that though these variables capture sources of marginalization, they are not exhaustive of all potential factors affecting patient experiences. Upon running the causal analysis, we removed any patients that had missing data for these features, which brought our patient count to 22,624. To err on the side of detecting more links from confounders, especially due to additional features and smaller number of data points, we increased the significance level to $\alpha = 0.15$, see Fig. \ref{fig:fci2}. Despite this, compared to Fig. \ref{fig:fci1}, only Gender became completely mediated through the new variables, while the direct connections from Age to Judgmental and Evidential, and the direct connection from Race to Stigmatizing persisted. Religion, interestingly, did not show direct influence on testimonial injustice. Marital status could potentially mediate Gender's influence and insurance type appears to mediate part of Race's and Gender's influence. However, the latter's influence on Injustice is complicated enough that the discovered causal link is direct rather than through any of the terms (despite the fact that Injustice itself is a function of those terms). There are possibly other mediators not measured in MIMIC-III that could affect the SCM. We conclude that although confounders may be present in general, the evidence MIMIC-III provides points to the causal influence of our chosen demographic variables on unjust terms and testimonial injustice to be rather direct.

\section{Intentional Interventions: A Use Case Study}
We conduct a study to understand how intentional interventions along the causal paths of our SCM (Fig \ref{fig:fci1}) can affect patient perception and ability to gauge their condition by human experts or an LLM (RQ2). We employ modifications to physicians notes by word removal, empathy reframing, and using preferred phrasing, with the goal of refining how patients are represented in such human and LLM interpretations. By clarifying marginalized patient experiences through precise linguistic adjustments, we take a step toward justice and better understand the potential of LLMs as collaborative partners in clinical practice. The development of patient-centered care heavily depends on the ability of sociotechnical systems to address these biases and ensure that every patient is truly understood. 

To further motivate edits, note that neutral text is the suggested way to discuss patient experiences in medical settings \citep{harrigian-etal-2023-characterization}. To address negative and stigmatizing language in text, alternative terms can be used to convey the same information in a more neutral manner. Additionally, to reduce evidential and judgmental terms, empathy can be injected through alternative empathic rephrasing or judgmental words can be removed entirely. We collect commonly used stigmatizing, judgmental, evidential, and negative terms in medical settings and preferred alternative phrasing for those terms \citep{beach2021testimonial, harrigian-etal-2023-characterization, ho2022stigmatizing, national2021words, judgealts}. The suggestions from these works, primarily from experts in the fields of substance use disorders and mental health, emphasize the following principles: avoiding overly positive reinterpretations of patient experiences, refraining from demonizing patients based on their condition, using empathy to describe the patient, adopting a neutral tone, and using first-person language such that a person's identity is detached from their condition.

With these interventions we see interesting insights which are discussed in the Results. This shows how even minimal modifications can lead to measurable changes in downstream outcomes. 

\paragraph{Editing Approaches. }We use the following editing approaches when making modifications to physicians' notes based on expert recommendations, as shown in Table \ref{table:samplealts} (full list in Table \ref{table:alts} of the Appendix):
\begin{itemize} \small
    \item \textbf{Word Removal} Adverbs modify verbs, adjectives, or other adverbs by conveying manner, conditions, or degree. In physicians' notes, they frequently add emphasis (e.g., blatantly), which can reinforce testimonial injustice. Accordingly, unjust terms ending in “-ly” were removed, as they were unnecessary and did not affect the meaning of the notes.
    \item \textbf{Empathy Reframing} Empathy, “acknowledging another’s emotional state without experiencing it” \citep{markakis1999teaching}, is critical in medicine for ensuring patients feel heard. Clinical empathy \citep{halpern2003clinical} uses reasoning to communicate effectively, unlike sympathy, which risks physicians assuming patients’ burdens. We applied clinical empathy to physicians’ notes by rephrasing content to detach patients from their conditions, a method effective in law \citep{lee2014judging} and medicine, while avoiding emotional language that could bias clinicians.

    \item \textbf{Preferred Rephrasing and Word Replacement} Synonyms are words or phrases that carry the same or similar meaning as another word. Many alternative wordings to judgmental and stigmatizing words have been proposed in clinical settings \citep{beach2021testimonial, harrigian-etal-2023-characterization, ho2022stigmatizing, national2021words, judgealts}. Here, we simply replace the words and/or phrasing in the text with select proposed terms by substance disorder and mental health experts. 
\end{itemize}

\begin{table} 
\centering
\resizebox{0.83\columnwidth}{!}{%
\centering
\begin{tabular} 
{|p{0.1\textwidth}|p{0.24\textwidth}|p{0.12\textwidth}|}
\hline
\textbf{Words} & \textbf{Alternatives} & \textbf{Term Type} \\ \hline
complains & expressed concerns about & Evidential \\ \hline
addict & has a use disorder & Stigmatizing \\ \hline
blatantly &  & Negative \\ \hline
apparently &  & Judgmental \\ \hline
\end{tabular}
}
\caption{Example of Alternative Words/Phrases to Testimonially Unjust Terms}
\label{table:samplealts}
\end{table}

\subsubsection{Logistic Regression} \label{ssec:logreg}

Our goal is to apply these edits \textit{only when they are needed}, in order to minimize the scale of changes. We do this based on the insights provided by the SCM. However, we need to distinguish instances of positive vs. negative causal influence. To achieve this, we train logistic regression models, 1 for each category of unjust term, to predict based on the 3 demographics and their conjunction whether they will incur a ``high'' occurrence (a value of 9).
We employ \texttt{scikit-learn} logistic regression models \citep{scikitlearn} to evaluate the causal relationships identified by our SCM (recall Fig. \ref{fig:fci1}). The coefficients of the resulting models are given  in Table \ref{table:logregresstable}.

The first thing to highlight is that the highest magnitude coefficients in each demographic column are indeed at the causal links identified by the SCM. Next, for each category of terms, we would like to build a ``simple'' rules based on the demographics, to characterize conditions under which we will experience more of those terms. Simplicity in this case is about using a single disjunction and/or conjunction. In some instances, such rules are easy to deduce. For example, \textbf{Race} is the most significant factor positively  influencing evidential terms, with \textbf{Gender} as the second most substantial contributor. Accordingly, we consider patients marginalized by \textbf{Race or Gender} for modifications to evidential terms. In other cases, it is less obvious how to distill such simple rules. So, instead, we built truth tables with the thresholded logistic regression predictions and identified the following rules, which disagree in at most 2 places with these truth tables:

\begin{itemize} \footnotesize
\itemsep-0.15em
    \item \textbf{Evidential:} {Race or Gender}
    \item \textbf{Judgmental:} {Race or Age or Gender} (Conjunction)
    \item \textbf{Negative:} {Gender or Age}
    \item \textbf{Stigmatizing:} {Race} and {(Gender or Age)}
\end{itemize}

\begin{table}[h]
\centering
\resizebox{\columnwidth}{!}{%
\begin{tabular}{|l|l|l|l|l|}
\hline
\textbf{Term Type}                              & \textbf{Race} & \textbf{Age} & \textbf{Gender} & \textbf{Conjunction} \\ \hline
\textbf{evidential terms (89.326\% accuracy)}   & 0.553 & -0.522 & -0.0347 & -0.2852           \\ \hline
\textbf{judgmental terms (90.183\% accuracy)}  & 0.621 & -0.383  & 0.0193 & 0.0184               \\ \hline
\textbf{negative terms (87.150\% accuracy)}     & 0.445 & -0.221 & -0.170 & -0.1999              \\ \hline
\textbf{stigmatizing terms (89.650\% accuracy)} & 0.638 & -0.222  & 0.0269 & 0.0645            \\ \hline
\end{tabular}
}
\caption{This table presents the logistic regression coefficients from experiments conducted on each term type (i.e., evidential, judgmental, negative, and stigmatizing) predicting whether a patient belongs to a marginalized demographic group (i.e., race, age, gender, or their conjunction, defined as having at least 1 marginalized feature ($is\_marginalized = 1$))}
\label{table:logregresstable}
\end{table}

\subsubsection{Rule-based: Causality-Informed Modifications} \label{ssec: causalinformededited}
In the rule-based approach, we perform string replacement using pattern matching to reframe and replace language (i.e., words and phrases) that leads to testimonial injustice. This is done with each of the aforementioned editing approaches: word removal, empathy reframing, and using preferred rephrasing. Our modifications were guided by the insights we gleaned from the SCM and logistic regression models. The causal graph analysis identified how patterns of testimonial injustice appear in the physicians' notes based on demographic features and their intersections (i.e., intersectionality), while logistic regression applied to the causal graph connections quantifies the strength of these relationships.

To trigger the word removal, empathy reframing, and preferred rephrasing modifications, we find if a patient is likely to be marginalized based on race, gender, age, or any combination of these features, as indicated by the variables: \textbf{is\_marginalized\_race = 1}, \textbf{is\_marginalized\_gender = 1}, \textbf{is\_marginalized\_age = 1}, and \textbf{is\_marginalized = 1}. The specific ways of the modifications are described in the Logistic Regression Section. An example of how a patient's notes are modified from the unedited version to the rule-based edits is illustrated in Fig. \ref{fig:mods}, the patient is an adult, Female Latina. Here we see how, for the rule-based approach, ``complains'' is updated to ``expressed concerns about''.

\subsubsection{Context-based: LLM Modifications} \label{ssec: llmedited}
\begin{figure}[h]
 \centering
 \includegraphics[scale=0.37]{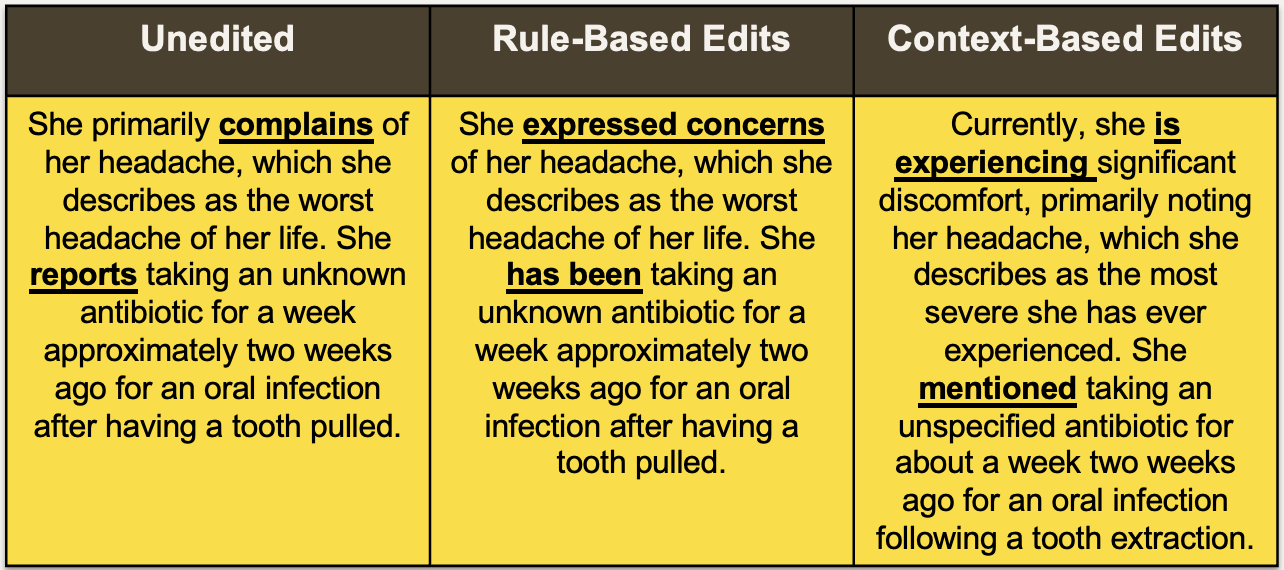}
 \caption{Example of Edited Notes Across Modification Types for an Adult, Female Latina}
 \label{fig:mods}
\end{figure}

In our context-based approach, we utilized OpenAI's GPT-4o-mini (paid tier 1) \citep{openai2024gpt4omini} to edit the EHR excerpts with specific instructions to act as ``an expert editor of clinician notes who wants to reduce negative and stigmatizing language in text using non-negative and non-stigmatizing language and reduce evidential and judgment, one can inject empathy/use alternative empathic rephrasing or remove words when possible.''  To achieve optimal performance, we carefully selected and fine-tuned the model’s parameters: \texttt{top\_p} = 1.0, \texttt{max\_tokens} = 700, and temperature = 0.5. We did not specify demographic features for the LLM; however, gender was frequently mentioned in the notes, and many conditions are age-related, making age easily inferrable by the model. More critically, the word choices in physicians' notes embed race-specific biases, enabling models to readily detect them \citep{gabriel2024can} (e.g., ``non-compliant'', ``resistant'' for Black patients). An example of how a patient's notes are modified from the unedited version to context-based edits is shown in Fig. \ref{fig:mods}, the patient is an adult, Female Latina. Here we see that, for the context-based approach, ``complains'' is updated to ``is experiencing''. Ironically, but not unexpectedly, we observed that many of the word replacements generated by the LLM during editing were also present in the lexicons and classified as testimonially unjust. The replacement choice ``reports'', or lack of replacement, in some instances, was among the most common.

\subsubsection{LLM Analysis of Notes Post-Modification} \label{ssec: llmanalysismethods}
To mitigate the risk of the model being biased against itself, especially after the context-based edits, we used a different LLM to analyze the notes. We asked Google DeepMind's Gemini 2.0 Experimental Flash \citep{google2024gemini2flash} (\texttt{top\_p} = 0.95 and temperature = 0.5) using the free-tier, the following questions to analyze each of the 24 physicians' notes. We instructed the model to ``answer the questions in a professional manner''. Here, we explore whether improved note-taking --- i.e., reduced testimonial injustice in the notes --- would enable an LLM to better perceive the patient or more accurately assess their condition.

\begin{itemize} \small
    \item How do you perceive the patient’s current condition? Select one of the following:
    \begin{itemize} \itemsep-0.25em
        \item Preventable due to better preventative care
        \item Preventable but caused by their own doing
        \item Likely environmental cause
        \item Unfounded cause
    \end{itemize}
    \item How do you perceive the patient? List four adjectives.
    \item What are the most critical symptoms of the patient?
    \item What are the least critical symptoms of the patient?
    \item What are the most likely diagnoses/conditions of the patient given what you have read?
    \item Should the patient receive immediate treatment or are they able to handle it at a later time? Select one: (Immediate, Later )
\end{itemize}

We ran the physicians' notes through Gemini asking the aforementioned questions without edits (i.e., as is directly from the \textsf{MIMIC-III} dataset), with edits as informed by the causal model, and with edits considering context by GPT-4o-mini. 

\subsubsection{Survey: Human Expert Analysis of Notes Post-Modifications} \label{ssec: humanmethods}
We surveyed US Board-Certified medical experts who regularly read physicians' notes and interact with patients --- specifically Physicians, Nurse Practitioners, and Physician Assistants to see if improved note-taking --- i.e., reduced testimonial injustice in the notes --- would enable them to better perceive the patient or more accurately assess their condition. They reviewed both edited and unedited versions of the notes and answered the same aforementioned questions. Additionally, we collected demographic information from them as listed in the Appendix.

We presented our human experts with 2 versions of the survey, determined by a coin-flip. If the participant reported they flipped ``Heads'' in their coin flip, they were presented with the first version of the survey, referred to hitherto as \textbf{VA}. Otherwise, they were presented with the second version of the survey, \textbf{VB}. VA contained the original, unedited notes from \textsf{MIMIC-III}, while VB included edited notes, evenly split between rule-based and context-based.

\subsection{Results} \label{ssec:interventionresults}
We hypothesize that testimonial injustice terms in ICU medical records can bias perceptions of patients and compromise fair assessment. Guided by our SCM, we adapt note excerpts to reduce injustice through word removal, alternative phrasing, and empathy reframing, aiming to promote more just and accurate patient evaluations.

\subsubsection{Semantic Similarity Post-Modifications}
One may wonder if such changes are actually changing the meaning of the patient summaries. Upon performing semantic similarity analysis (cosine similarity \cite{reimers-2019-sentence-bert}) on the embeddings of the experts of the patient summaries for rule-based and context-based edits, we found the minimum similarity values were high across both types with senior Latino males having 97.63\% and adult Latino males having 86.11\% respectively. Maximum similarity for child White males at 100\% for rule-based edits and adult White males at 99.1\% for context-based edits.

\begin{figure*}[h]
 \centering
 \includegraphics[scale=0.57]{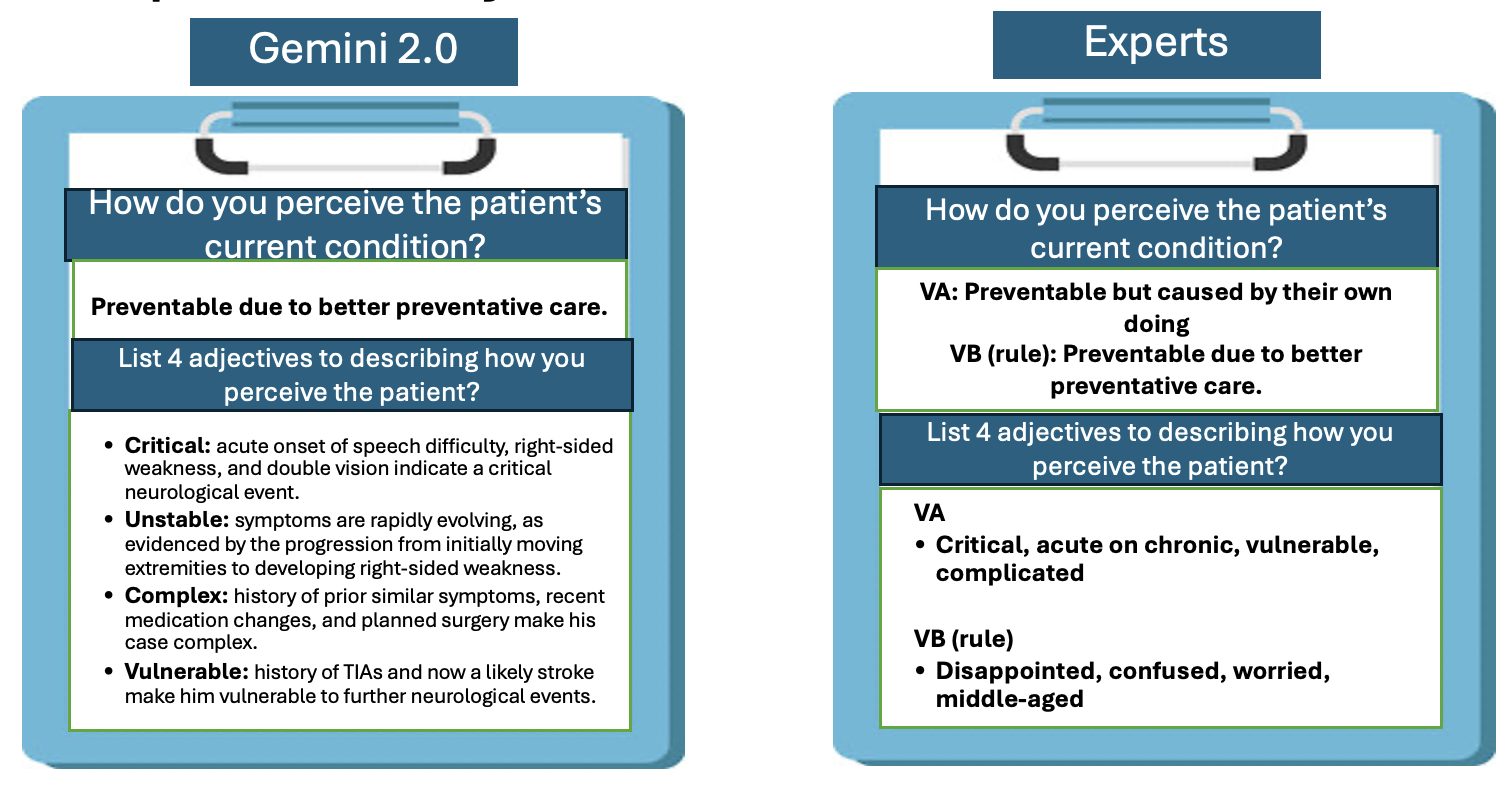}
 \caption{Example of Expert and Gemini 2.0 Feedback on Patient Notes Post-Modifications}
 \label{fig:feed}
\end{figure*}

\subsubsection{LLM Analysis}
\begin{figure}[th]
 \centering
 \includegraphics[scale=0.25]{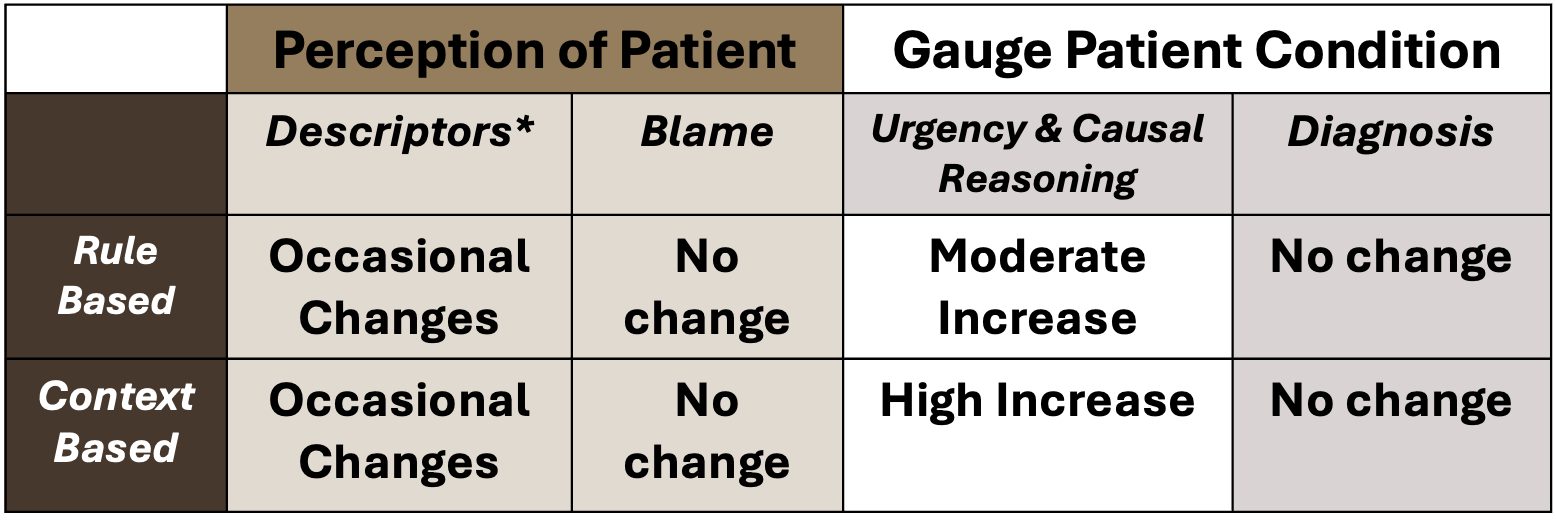}
 \caption{Trends Post-Modifications of Notes in Rule-Based and Context-Based Approaches Across Patient Perception and Condition by Gemini 2.0 Flash. *LLMs have little variety in describing patients and their experiences}
 \label{fig:genllm}
\end{figure}

We sent the 3 versions of the 24 notes to Gemini: (1) unedited, (2) rule-based edits, and (3) context-based edits. The general trend changes in Gemini's responses across patient perception and gauging the condition of the patient, post-modifications, are summarized in Fig. \ref{fig:genllm}. Both approaches show occasional changes in patient descriptors, but no change in blame or diagnosis. However, urgency \& causal reasoning increased in the rule-based approach and even more significantly in the context-based approach.

For the unedited notes, the model's responses were efficient and generally neutral. However, after editing the notes, we observed the greatest variability in the model's justification when answering the question of whether a patient's symptoms required immediate or delayed attention. Notably, urgency increased with rule-based edits and, at times, even more so with context-based edits. Additionally, for the same question, both rule-based and context-based edits produced more detailed justifications, linking symptoms to potential conditions or diseases in a causal manner. Context-based edits further emphasized the urgency of the issue, whereas unedited text often led to more direct recommendations. Lastly, we saw after the edits only 1 instance of variability when the model was asked to give adjectives of the patient whereas in the unedited version the list of adjectives changed from Critical, \emph{Unstable}, Compromised, High-Risk to Critical, \emph{Vulnerable}, Compromised, High-Risk post edits. By observation, in this medical setting, ``unstable'' was mostly used to describe patients whose conditions could change at any moment. While ``vulnerable'' was most used to describe patients who were likely to contract another disease or have worsening conditions without immediate treatment. It follows logically that vulnerability can be seen as a more specific characterization of unstable in this context. This change only occurred once in this very small sample set. However, with a larger dataset, it is reasonable to expect that changes in how patients are described (through adjectives) could vary with edits—just as the perceived urgency of their conditions changed. It is important to recall that these physicians' notes originate from ICU settings, where the needs of every patient are always immediate. However, adding context, detailed explanations, and a sense of urgency can enhance understanding and improve the perception of the patients' conditions. Therefore, when LLMs are fed physicians' notes, physicians can aid the perception of their patients and clarity in the actual condition by being mindful of their word choice and, when appropriate, editing their notes. An example of the feedback provided by Gemini 2.0 Flash is shown in Fig. \ref{fig:feed}.

A limitation of the model’s responses was its tendency to interpret several aspects of a patient’s experiences as symptoms, despite categorizing them as the least critical. For instance, in the case of a adult Black woman, the original medical record stated: ``The patient lives alone in the hospital. She is currently unemployed.'' GPT-4.o-mini revised this to ``The patient is currently hospitalized and lives independently'', whereas the rule-based edits left the text unchanged due to none of the terms being present in the lexicons of concern. Here, Gemini identified ``Lives Alone and Unemployed'' as 1 of the least critical symptoms. Additionally, the model’s responses lacked diversity in the adjectives used to describe patients. This is likely due to the tendency of LLMs to over-generalize \citep{allaway2024exceptions}. 

\subsubsection{Human Expert Analysis}
We presented our human experts with 2 versions of the survey, VA and VB, as described previously.
An example of the feedback provided by the experts is shown in Fig. \ref{fig:feed}.  We had 17 complete expert surveys, with each expert reviewing 5 distinct patient cases, resulting in 85 expert evaluations. Thus, this experimental design provided multiple independent judgments per condition, increasing the robustness and reliability of our findings.

To give more insight, surveys that were not completed typically ended at the coin-flip question. Since we provided a link to an online coin flipper, the lack of a physical coin was not a limiting factor. However, the survey length may have been a challenge, as it took participants an average of 25 minutes to complete. Also, there were slightly more respondents who flipped heads (3) than tails --- well within the typical statistics for a Bernoulli($\frac{1}{2})$ random variable.

\begin{figure}[th]
 \centering
 \includegraphics[scale=0.25]{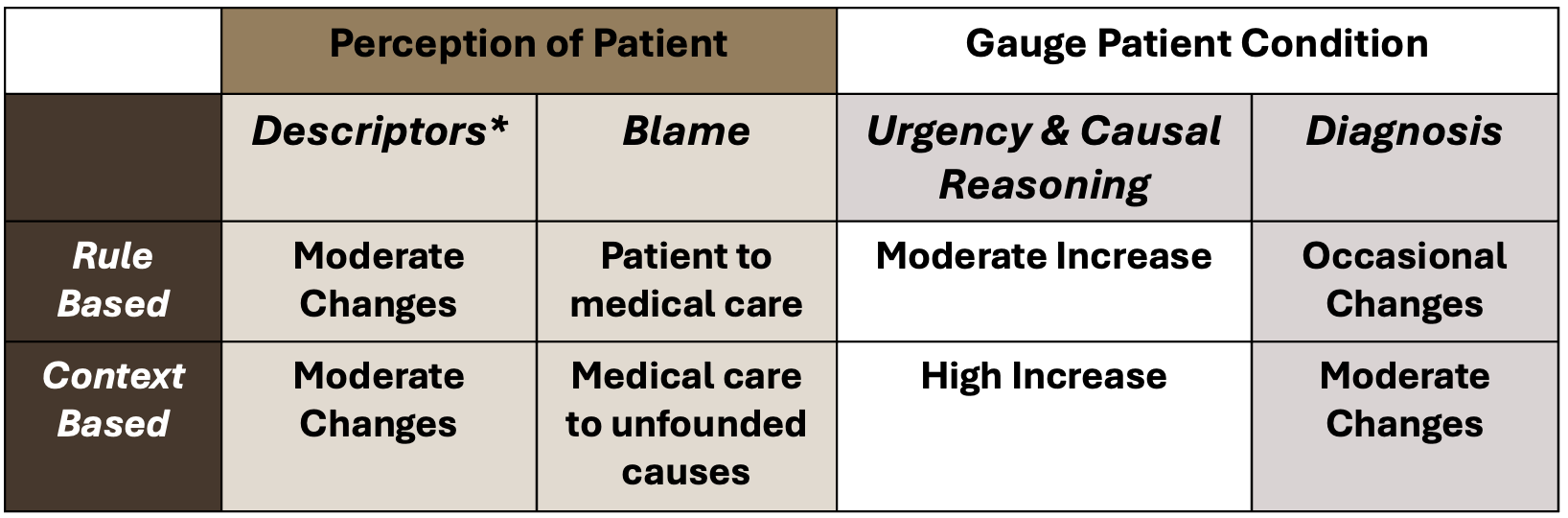}
 \caption{Trends Post-Modifications of Notes in Rule-Based and Context-Based Approaches Across Patient Perception and Condition by Experts. *Humans had various ways of describing patient experiences due to their vast vocabulary.}
 \label{fig:genexpert}
\end{figure}

Each participant reviewed 5 physicians' notes in VA or VB, randomly selected from the set of 24 patients to improve response quality and manage survey length.
All but 3 patient profiles were randomly selected for review and none were reviewed more than 3 times. The 3 profiles that were not randomly selected were: adult White female, child Black female, and child Asian female. With a larger sample size and extended survey time, all patient profiles would likely have been reviewed at least once. The general trend changes in human experts' opinions across patient perception and gauging the condition of the patient are summarized in Fig. \ref{fig:genexpert}. There were frequent changes in patient descriptors across both methods, as the clinicians had a number of different descriptors chose to use (since this was an open ended question in the survey). It primarily reflects the richness of their vocabularies. The blame for the patient's condition tended to shift from the patient to poor medical care in the rule-based approach and to unfounded causes in the context-based approach. Urgency \& causal reasoning increased in the rule-based approach and even more significantly in the context-based approach, just as with the LLM analysis. Diagnosis changed occasionally in the rule-based approach and more frequently in the context-based approach.

When asked, “How do you perceive the patient’s current condition?”, human experts at times attributed conditions to ``unfounded causes'' even in unedited notes (VA), which did not occur for the notes analyzed with Gemini. Further, they tended to shift blame away from patients and toward poor medical care. In cases where experts blamed the patient for their condition in unedited notes, rule-based edits most often led them to shift blame toward poor medical care. Overall, rule-based edits resulted in the least blame attributed to patients, with more blame assigned to unfounded or environmental causes. However, context-based edits reversed this trend, leading to an increase in patient blame and a slight increase in unfounded causes (just 1 additional case).

We also observed high variability in expert responses to the same question, particularly regarding the cause of the patient’s condition. Across the 21 profiles, respondents disagreed on 16 cases but had alignment with 5 of the VA excerpts and 4 with VB. Additionally, expert analysis revealed that some patients were perceived as eligible for delayed treatment rather than immediate care. This occurred only once, with responses remaining consistent across the 2 participants who answered this question --- 1 reviewing VA and the other reviewing VB with context-based edits. 

We do see that physicians gauge conditions differently for 3 patients under context-based edits and 1 patient under rule-based edits. Specifically, under context-based, pneumonia was newly perceived as hypertension, heart failure or edema as an infected AV site, which can lead to edema, and gastrointestinal issues and ulcers as COPD. After rule-based edits, a urinary tract infection was reclassified as a stroke.

Finally, human experts used a wider variety of adjectives to describe patients and their experiences compared to the LLM. Unlike the LLM, experts did not misinterpret life experiences as symptoms.


\subsubsection{Sentiment Analysis}
We use the nltk library's SentimentIntensityAnalyzer \citep{nltklib} to perform sentiment analysis on the excerpts of physicians' notes and feedback from Gemini and the human experts. 

\noindent\emph{\textbf{Patient Adjectives by LLM: } }Across all 24 patients, each has at least 2 negative adjectives, and 22 have at least 1 neutral adjective, while 2 have a positive adjective. The most common negative adjectives are Unstable'', Vulnerable'', and Critical'', with Critical'' reflecting life-threatening ICU conditions in the \textsf{MIMIC-III} dataset. Neutral adjectives, such as Complex'' and Monitored'', describe multifaceted conditions or observation status. The only positive adjective, ``Challenging'', likely carries negative connotations due to patient apprehension. Sentiment analysis across original, rule-based, and LLM-edited responses shows negative sentiment dominates in all categories, with 24 negative and 4 neutral responses for original and LLM-edited responses, and 23 negative and 5 neutral for rule-based edits. Overall, LLM-generated descriptions predominantly use negative terms, with few neutral and almost no positive descriptors. 

\noindent\emph{\textbf{Patient Adjectives by Experts:}} Experts generally provided neutral responses overall, though their adjectives reveal more nuance. For VA, all but 2 patients had at least 2 neutral adjectives, with the remaining patients described solely with negative terms. Common negative adjectives were Ill'', Vulnerable'', and Confused''; neutral adjectives included Complex'', Guarded'', and Treatable''. Positive adjectives appeared once (Determined''). For VB (context-based), frequent negative terms were Anxious'', Ill'', Grimacing'', and Frustrated'', while neutral adjectives included Tachypneic'', Edematous'', and Resilient''; positive terms appeared once (``Hopeful''). Overall, neutral terms were most frequent, consistent with recommended documentation practices. Term counts were: VA—31 neutral, 27 negative, 6 positive; VB (context) —32 neutral, 24 negative, 3 positive; VB (rule)—14 neutral, 11 negative, 2 positive.

\noindent\emph{\textbf{Edits}}
The sentiment analysis of the unedited excerpts, rule-based edits, and LLM-based edits of the EHRs revealed some differences in how sentiment was altered across these versions. The original excerpts showed a predominance of negative sentiment, with 21 instances categorized as negative, 5 neutral, and 2 positive. Upon editing EHRs according to the causal-informed rule-based edits, the number of negative excerpts decreased slightly to 19, while the neutral excerpts increased to 6, and the positive rose to 3. This indicates that the rule-based edits softened the negative tone slightly but did not drastically change the overall sentiment distribution. The LLM-based edits showed a similar decrease in negative sentiment, with 19 instances, however, the number of positive excerpts increased more significantly, from 3 in the rule-based edits to 5, and the neutral category slightly decreased from 6 in the rule-based edits to 4. This suggests that the LLM-based edits introduced a more positive tone, compared to both the original and rule-based edits. Overall, both rule-based edits and LLM-based edits reduced the negative sentiment across the notes compared to the unedited versions. The LLM-based edits led to a more pronounced shift towards positive sentiment, while the rule-based edits had a more modest effect which is a much more desirable trait in medical writing and the preferred style of speech in medicine \citep{healy2022reduce}.

\section{Discussion and Conclusion} \label{sec:discuss}
Our findings demonstrate that age, gender, and race each contribute to testimonial injustice in distinct and interacting ways, with evidential, judgmental, and stigmatizing terms driving most unjust documentation. Race is associated with stigmatizing and judgmental language, suggesting that patients’ experiences—particularly in high-stakes contexts like the ICU—are often diminished or vilified. Age influences the use of evidential and judgmental terms, indicating that even when patients are acknowledged, physicians may remain skeptical or critical. Gender is linked to negative terms, contributing to experiences being dismissed. Negative terms rarely appear in isolation, instead emerging alongside other aspects of injustice. These findings highlight the importance of intersectionality: no single factor explains unjust language, and understanding their interactions is essential to target interventions that can mitigate these harms. Ethically, these disparities represent not only misperception but a failure to uphold equitable, just care, with implications for patient trust and wellbeing.

Our causal analysis situates these patterns within a broader institutional and societal context. Although our data focus on ICU notes, similar cognitive and systemic biases are pervasive across healthcare, as noted by the American Medical Association. The disparities we observe are not merely descriptive; they reveal mechanisms through which structural and individual biases are encoded in clinical documentation. Interventions that address these patterns—through careful language choices, rule-based or context-informed edits, or awareness training—can influence how both human experts and AI systems perceive patients, helping to reduce misdiagnosis, unjust blame, and inequitable outcomes. At the same time, these interventions must be implemented thoughtfully, acknowledging potential limitations, such as over-correction, clinician resistance, or misinterpretation by downstream AI tools.

Importantly, both human experts and LLMs responded positively to edited notes: LLMs expressed greater urgency, provided more contextualized reasoning, and linked symptoms to severity more accurately, while human experts shifted blame away from patients toward systemic factors, particularly under rule-based edits. These results underscore that small, targeted interventions in documentation can have measurable downstream effects on perception and decision-making. Moving forward, encouraging awareness of testimonial injustice, promoting unbiased documentation, and integrating these insights into clinical AI systems are critical steps toward equitable, ethical, and effective healthcare assessment.

\section{Limitations \& Future Work}
\noindent\textbf{Better data:} Boston has a median age of approximately 33 for both males and females, with a racial composition of White (50.13\%), Black (21.7\%), Asian (9.59\%), and Latino (5.92\%) \citep{bostoncensus}. While race emerges as a frequent contributor to testimonial injustice, age and gender also play important roles, reflecting nuanced intersectional effects. More representative datasets of marginalized groups are critical to advance this work, yet such data are rare due to limited access to care and historical distrust in healthcare systems. Expanding datasets to include balanced race, age, and gender distributions would improve causal inference and allow incorporation of features known to clinicians but not otherwise observable. Examination of our SCMs further reveals potential unknown confounders; for instance, stigmatizing and negative terms may co-occur due to implicit biases, potentially linked to unobserved, inferrable characteristics such as experiences specific to women (e.g., perceived obesity). These findings underscore the need to account for intersectional identities and systemic factors to fully understand and mitigate testimonial injustice. \\

\noindent\textbf{Analysis:} In conducting further analysis on the expert perceptions, we would like to understand the nuances that occur across them, given their own features. Specifically, we believe it worth accounting for personal features such as length of practice, sex, and the primary medical institution in terms of how they influence their assessments. 

\section{Ethical Considerations Statement}
In this work, we examine whether mitigation of harms in clinical language can meaningfully benefit populations historically marginalized within ICU and broader healthcare settings, as defined by prior literature. While improvements in language and clinical modeling may provide broader benefits across patient populations, our primary objective in this work is to investigate whether such approaches can improve representation, accessibility, and care-related outcomes for groups that have historically experienced disproportionate barriers to equitable and accurate treatment and participation in healthcare systems. This focus is intentional rather than exclusionary: we do not regard harms experienced by other populations as unimportant, but instead prioritize evaluating if these methods can address disparities affecting communities that are disproportionately impacted by structural inequities, experience systemic barriers to care, or are historically excluded not only within clinical settings, but across other high-stakes contexts as well. These considerations highlight the importance of design choices in such settings, often sensitive to structural inequities, and their downstream effects on patients of marginalized backgrounds.

\section{Acknowledgments}
This work was partially supported by the NSF Program on Fairness in AI in Collaboration with Amazon under Award No. IIS-1939743, titled FAI: Addressing the 3D Challenges for Data-Driven Fairness: Deficiency, Dynamics, and Disagreement. Any opinion, findings, and conclusions or recommendations expressed in this paper are those of the authors and do not necessarily reflect the views of the National Science Foundation or Amazon.




\bibliography{aaai2026}

\onecolumn
\appendix
\section{Appendix}

\subsection{Algorithmic Variation: PC} 
\begin{figure}[h]
    \centering
    \begin{subfigure}[t]{0.4\textwidth}
        \centering
        \includegraphics[height=1.75in]{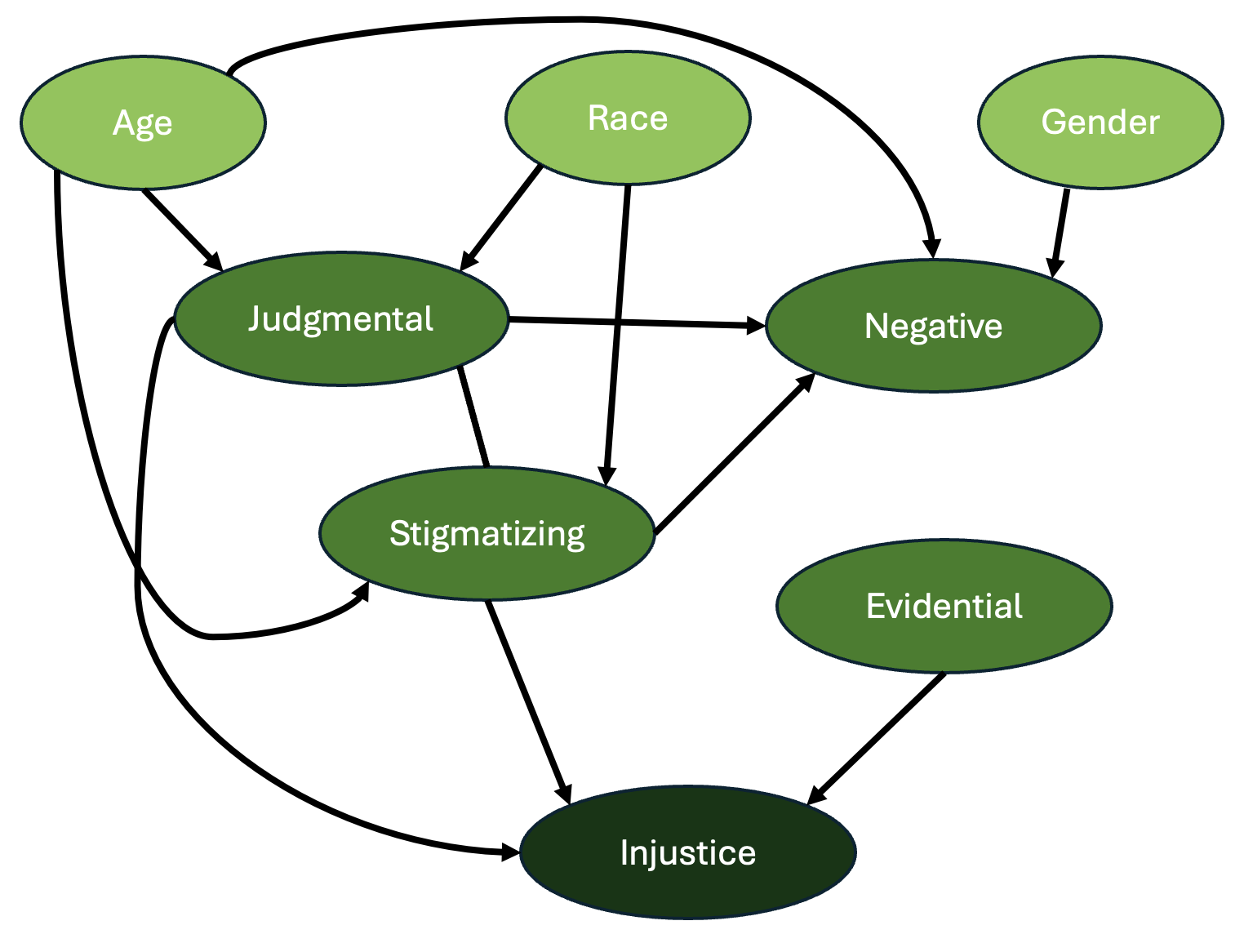}
        \caption{$\alpha = 0.01$}
        \label{fig:pc_001}
    \end{subfigure}
    \hfill
    \begin{subfigure}[t]{0.4\textwidth}
        \centering
        \includegraphics[height=1.75in]{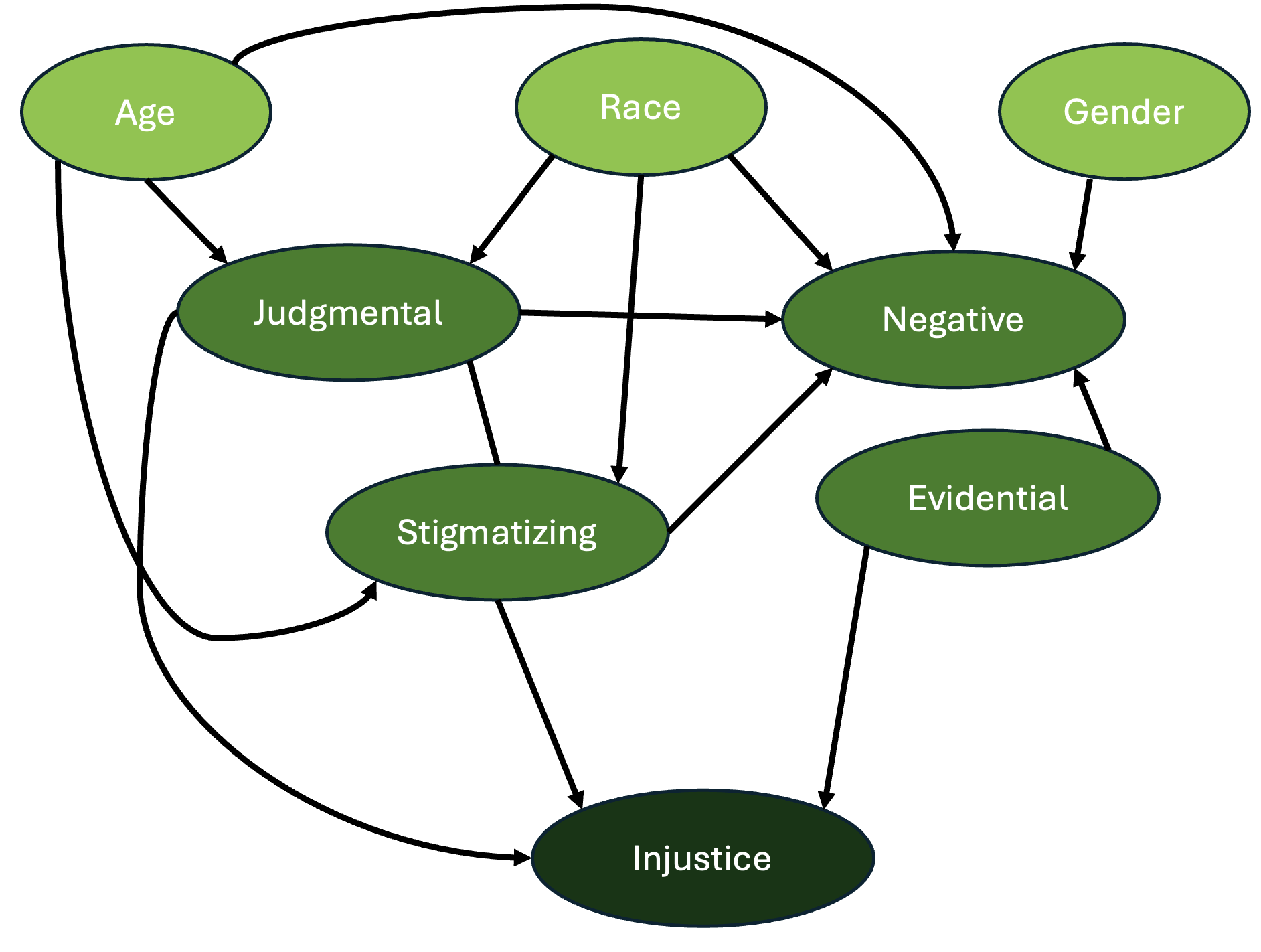}
        \caption{$\alpha = 0.05$}
        \label{fig:pc_005}
    \end{subfigure}
    \hfill
    \begin{subfigure}[t]{0.3\textwidth}
        \centering
        \includegraphics[height=1.75in]{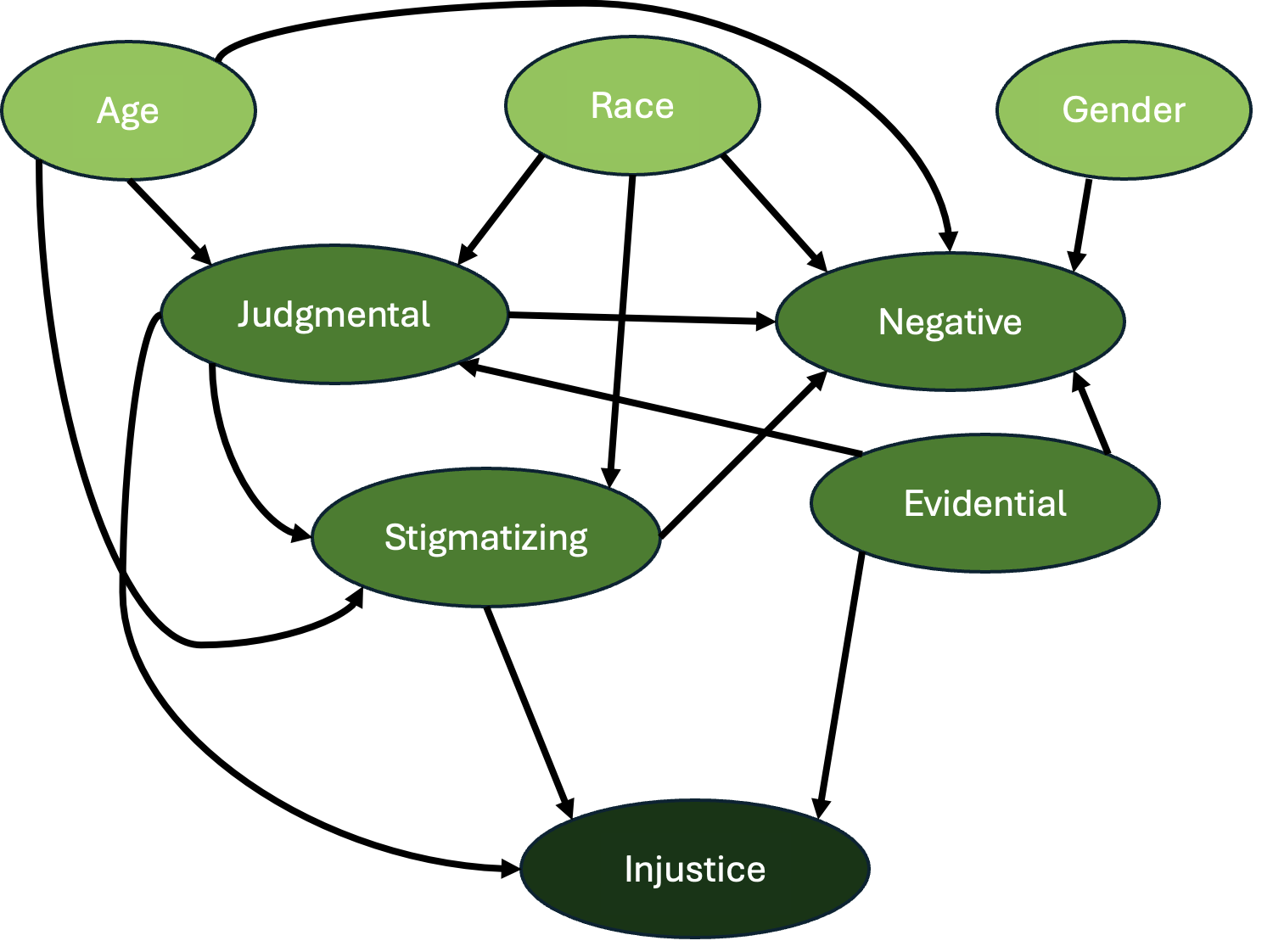}
        \caption{$\alpha = 0.1$}
        \label{fig:pc_01}
    \end{subfigure}
    \caption{Structural causal models (SCMs) inferred by PC at multiple significance levels ($\alpha = (a) 0.01, (b) 0.05, (c) 0.1$), highlighting the sensitivity of discovered dependencies and edge orientations to the choice of threshold.}
    \label{fig:pc_appendix}
\end{figure}

The Peter-Clark (\textsf{PC}) Algorithm \cite{spirtes2000causation} is a traditional constraint-based causal discovery algorithm that uses conditional independence testing to form causal relationships. The assumptions of the \textsf{PC} algorithm are that the true graphs follow the Markov Condition, Faithfulness Condition, and Causal sufficiency (discussed in the Assumptions section). We used \textsf{PC} to see how much the SCMs are affected by algorithmic variations. \textsf{PC} yields similar results to the \textsf{FCI} experiments, under the same experimental runs with similar $\alpha$-values. We show these in Fig. \ref{fig:pc_appendix}. The common trends that occurs across FCI and PC algorithms are that Race is a cause of both judgmental and stigmatizing terms, gender continues to be the sole root (i.e., demographic feature) contributor of negative terms, and evidential and stigmatizing terms are always direct contributors to someone experiencing testimonial injustice.

\clearpage
\subsection{Fast Causal Inference (\textsf{FCI}) } \label{ssec: FCIMethods}
Fast Causal Inference \textsf{FCI} \citep{spirtes2000causation} is a constraint-based causal discovery algorithm that takes as input our data and produces a Structural Causal Model (SCM) that captures the causal influences between all variables. (We employ the \texttt{causal-learn} library's Python implementation \citep{causallearn}.) The advantage of this algorithm over the typical \textsf{PC} Algorithm (discussed further in the Appendix), is that it allows for unknown confounders. This is particularly useful in this medical setting where we are only considering demographic features that can be observationally inferred. For instance, the specific condition one has could also have a large effect on the views of the doctors towards them, i.e., preventable diseases tend to carry more negative and stigmatizing language to inherited conditions \citep{beach2021testimonial, stig-effects}. \textsf{FCI} is also more informative about confounders and potential directions of causation than \textsf{PC}. The assumptions of the \textsf{FCI} algorithm are that the true graphs follow the Markov and Faithfulness conditions. \textbf{Markov Condition} is met if and only if a node, given its set of parents, is probabilistically independent of all of its children nodes in a graph. \textbf{Faithfulness Condition} is met if and only if there is no conditional independence in the graph that is not entailed by the Markov Condition. The goal of obtaining  a causal model is to elucidate what intersectional interactions give rise to what categories of unjust term, with the hope of subsequently using these insights for precise interventions.

We slightly vary $\alpha$-values to further our exploratory analysis of Structural causal models (SCMs) inferred by FCI at significance levels ($\alpha = (a) 0.01, (b) 0.05, (c) 0.1$).

\begin{figure}[h]
    \centering
    \begin{subfigure}[t]{0.4\textwidth}
        \centering
        \includegraphics[height=1.75in]{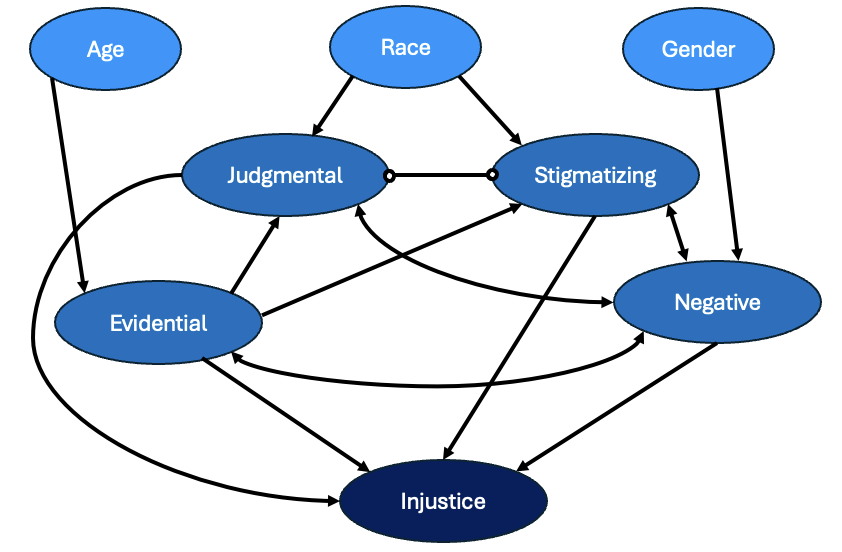}
        \caption{$\alpha = 0.01$}
        \label{fig:fci_001}
    \end{subfigure}
    \hfill
    \begin{subfigure}[t]{0.4\textwidth}
        \centering
        \includegraphics[height=1.75in]{figures/FCI_0.05_CRL.png}
        \caption{$\alpha = 0.05$}
        \label{fig:fci_005}
    \end{subfigure}
    \hfill
    \begin{subfigure}[t]{0.3\textwidth}
        \centering
        \includegraphics[height=1.75in]{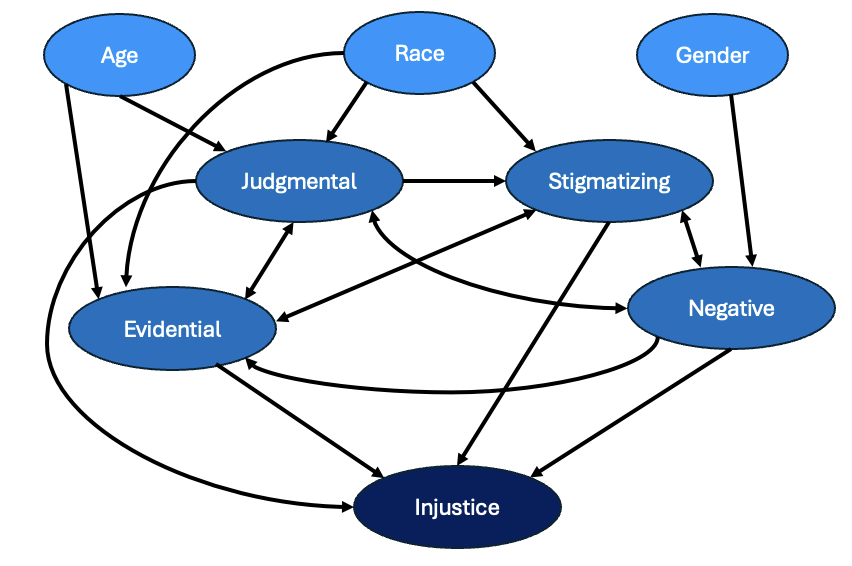}
        \caption{$\alpha = 0.1$}
        \label{fig:fci_01}
    \end{subfigure}
    \caption{Structural causal models (SCMs) inferred by FCI at multiple significance levels ($\alpha = (a) 0.01, (b) 0.05, (c) 0.1$), highlighting the sensitivity of discovered dependencies and edge orientations to the choice of threshold.}
    \label{fig:fci_appendix}
\end{figure}

\subsection{Importance of Intersectionality}
\begin{figure}[h]
\includegraphics[width=4cm]{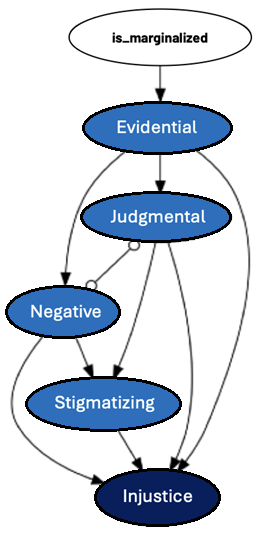}
\centering
\caption{FCI SCM with coarse granularity,  $\alpha = 0.05$}
\label{fig:fci_coarse}
\end{figure}

What would we miss if we do not take intersectionality into account? In the ML literature, it is more common to binarize commonly marginalized attributes of people. To show that a lot of nuance is lost, we now coarsen our marginalization feature to a single binary one, and show the effects of doing so in causal discovery. The results are given in Fig. \ref{fig:fci_coarse}. More precisely, we combine and equalize race, gender, and age such that if any patient is likely to experience marginalization along any of these lines, then they are considered to experience marginalization, i.e., if \texttt{\footnotesize is\_marginalized\_gender = 1} or  \texttt{\footnotesize is\_marginalized\_race = 1} or \texttt{\footnotesize is\_marginalized\_age = 1} then \texttt{\footnotesize is\_marginalized = 1}. We are unable to determine which specific features contribute to the experience of certain terms. We do still see that evidential terms, which dismiss individuals, is an entry point to them experiencing testimonial injustice. However, we lose any insight about the nuanced pathways to which race, age, and gender are contributors of testimonial injustice.

\subsection{Lexicon of Testimonial Unjust Terms}
Certain terms may be considered normative within medical settings and are often assumed to have no effect on patients. However, our work suggests that despite the 'norm', the use of words that contribute to testimonial injustice should be reconsidered. The presence or absence of these words shapes the narrative constructed around a patient and their experiences. Such a consideration becomes increasingly important as we rely more and more on AI-enabled tools to process and interpret clinical notes. These systems generate outputs based on statistical patterns in language, without an inherent understanding of clinical context, social meaning, or lived experiences. As a result, already ignored shortcomings may become amplified in these life-critical settings. We believe that strategic interventions in how clinical language is documented and processed could help mitigate these risks. Here, we list the lexicon of terms used to detect testimonial injustice: 
\begin{itemize}
    \item evidentials:
    \begin{itemize}
        \item  ``complains'',  ``denies'',  ``endorses'',  ``notes'',  ``reports'',  ``says'',  ``tells me''
    \end{itemize}
    \item judgmentals:
    \begin{itemize}
        \item  ``adamant'',  ``apparently'',  ``claims'',  ``insists'',  ``states''
    \end{itemize}
    \item negatives:
    \begin{itemize}
        \item  ``challenging'',  ``combative'',  ``defensive'',  ``exaggerates'',  ``disagreeably'',  ``deceitfully'',  ``deceptively'',  ``non'',  ``blatantly'',  ``absurdly'',  ``alarmingly''
    \end{itemize}
    \item stigmatizing:
    \begin{itemize}
        \item  ``cheat',  ``non-adherent',  ``refuse',  ``unwilling',  ``user',  ``adherence',  ``uncontrolled', 
  ``malinger',  ``pill problem',  ``non-compliant',  ``non-compliant'',  ``narcotics',  ``drug problem', 
  ``pill seeking',  ``in denial',  ``junkie',  ``been clean',  ``unmotivated',  ``fails',  ``cheats',  ``narcotic',  ``non-adherence',  ``faking',  ``combative',  ``failure',  ``argumentative'',  ``degenerate', 
  ``abuser'',  ``adherent'',  ``addicted'',  ``compliant'',  ``lifestyle disease'',  ``controlled'',  ``addict'', 
  ``fail'',  ``secondary gain'',  ``abuse'',  ``substance abuse'',  ``malingers'',  ``failed'',  ``controls'', 
  ``difficult patient'',  ``speed ball'',  ``drug seeking'',  ``strung out'',  ``abusing'',  ``malingerer'',  ``abuses'', 
  ``pot head'',  ``malingering'',  ``refuses'',  ``belligerent'',  ``fake'',  ``habit'',  ``alcohol abuse'',  ``compliance'', 
  ``control'',  ``refused'',  ``depraved'',  ``cheating''
        
    \end{itemize}
\end{itemize}


\subsection{LLM Feature Justifications}
The \texttt{top\_p} parameter, also known as nucleus sampling, determines the cumulative probability threshold for selecting the next token. By setting \texttt{top\_p} to 1.0, we allowed the model to consider all possible tokens without restriction, ensuring maximum flexibility for edits. This was important for our task, as limiting the token pool prematurely (e.g., with a lower \texttt{top\_p} value) could risk nuanced rephrasing options needed for the context of the EHRs. The temperature parameter controls the randomness of token selection, with lower values producing more deterministic and focused outputs, while higher values yielding more creative and varied responses. We set the temperature to 0.5 as a midpoint to encourage the model to generate thoughtful edits that adhered to the task’s constraints while avoiding overly rigid or excessively creative outputs. This balance ensured that the language remained professional, empathetic, and suitable for a clinical setting. A lower temperature (e.g., 0.1) might have resulted in overly repetitive or conservative edits. The \texttt{max\_tokens} parameter was set to 700 to control the length of the generated edits. This limit ensured that the modified excerpts remained approximately the same length as the original text, preserving the clinical document’s overall structure and readability, as excessively long outputs might introduce redundancy, while overly short edits risked omitting essential information. Together, these parameters were chosen to optimize the model’s ability to perform precise, context-aware edits while maintaining the integrity and usability of the original EHR excerpts.

\subsection{Experiences of Unjust Terms}

We report both the absolute and per-patient counts of unjust terms experienced by patients across demographic groups (race, gender, and age) within each category of testimonial injustice. Specifically, we examine evidential, judgmental, negative, and stigmatizing terms to characterize the distribution and intensity of injustice experienced by individual patients and demographic subgroups. This approach allows us to capture not only the total burden of unjust terms but also the per-patient exposure, highlighting potential disparities across marginalized populations.

\begin{table}[h]
\fontsize{8pt}{9pt}\selectfont
\centering
\begin{tabular}{l|l|l|l|l|l|l|l|l|l|l|}
\cline{4-11}
\multicolumn{3}{c}{} & \multicolumn{4}{|c|}{Overall Word Counts} & \multicolumn{4}{|c|}{Per Patient Word Counts } \\ \hline
\textbf{Race}   & \textbf{Gender} & \textbf{Age}     & {evidential}  & {judgmental}  & {negative}    & {stigmatizing}  & {evidential}  & {judgmental}  & {negative}   & {stigmatizing} \\ \hline
Asian  & Female & Senior & 17787  & 1810  & 13881  & 9094   & 80   & 8   & 63  & 47  \\ \hline
Asian  & Female & Adult  & 15873  & 1624  & 12453  & 9354   & 100  & 11  & 68  & 50  \\ \hline
Asian  & Female & Child  & 10166  & 1087  & 6959   & 5118   & 92   & 10  & 70  & 42  \\ \hline
Asian  & Male   & Senior & 27854  & 2927  & 21341  & 12801  & 97   & 10  & 74  & 51  \\ \hline
Asian  & Male   & Adult  & 26017  & 2623  & 19855  & 13496  & 136  & 16  & 143 & 108 \\ \hline
Asian  & Male   & Child  & 16186  & 1856  & 16978  & 12909  & 165  & 18  & 132 & 107 \\ \hline
Black  & Female & Senior & 156312 & 16820 & 124588 & 101201 & 162  & 18  & 136 & 109 \\ \hline
Black  & Female & Adult  & 177194 & 20246 & 148941 & 119566 & 180  & 21  & 150 & 126 \\ \hline
Black  & Female & Child  & 86630  & 10054 & 72087  & 60767  & 170  & 19  & 137 & 115 \\ \hline
Black  & Male   & Senior & 132160 & 14430 & 106435 & 89615  & 202  & 24  & 159 & 126 \\ \hline
Black  & Male   & Adult  & 177149 & 20942 & 139331 & 109909 & 171  & 18  & 128 & 105 \\ \hline
Black  & Male   & Child  & 66758  & 7023  & 49991  & 40998  & 132  & 11  & 107 & 84  \\ \hline
Latino & Female & Senior & 10681  & 901   & 8687   & 6776   & 150  & 14  & 131 & 130 \\ \hline
Latino & Female & Adult  & 8404   & 797   & 7327   & 7291   & 210  & 17  & 198 & 240 \\ \hline
Latino & Female & Child  & 5666   & 468   & 5359   & 6473   & 112  & 12  & 79  & 62  \\ \hline
Latino & Male   & Senior & 9757   & 1072  & 6910   & 5416   & 161  & 16  & 123 & 112 \\ \hline
Latino & Male   & Adult  & 17561  & 1755  & 13429  & 12247  & 229  & 19  & 171 & 187 \\ \hline
Latino & Male   & Child  & 10298  & 856   & 7697   & 8437   & 103  & 11  & 76  & 61  \\ \hline
White  & Female & Senior & 625318 & 64160 & 464391 & 370000 & 97   & 10  & 71  & 56  \\ \hline
White  & Female & Adult  & 628096 & 63568 & 461312 & 363150 & 102  & 10  & 74  & 58  \\ \hline
White  & Female & Child  & 293618 & 29498 & 213835 & 165485 & 102  & 11  & 76  & 61  \\ \hline
White  & Male   & Senior & 830821 & 87542 & 619015 & 497329 & 106  & 11  & 77  & 64  \\ \hline
White  & Male   & Adult  & 896419 & 93996 & 650824 & 542044 & 103  & 11  & 75  & 63  \\ \hline
White  & Male   & Child  & 384385 & 40112 & 278847 & 233649 & 1131 & 118 & 820 & 687 \\ \hline
\end{tabular}
\caption{Absolute and per-patient numbers of unjust terms experienced by patients --- by race, gender, and age --- in each category of unjust terms leading to testimonial injustice --- evidential, judgmental, negative, and stigmatizing.}\label{tab:percdistribution}
\end{table}

\clearpage
\subsection{Alternate Words/Phrases}
\begin{table}[h]
\centering
\begin{tabular}{|p{0.27\textwidth}|p{0.3\textwidth}|p{0.14\textwidth}|}
\hline
\textbf{Original Words/Phrases} & \textbf{Alternatives} & \textbf{Term Type} \\ \hline
addict & has a use disorder & Stigmatizing \\ \hline
addicted & has an addictive disorder & Stigmatizing \\ \hline
alcoholic & has alcoholism & Stigmatizing \\ \hline
abuse & use disorder & Stigmatizing \\ \hline
refuse & discussed alternatives to & Stigmatizing \\ \hline
non-compliant & has a different care preference & Stigmatizing \\ \hline
the scourge of addiction & has a substance use disorder & Stigmatizing \\ \hline
addictive personality & has an addiction & Stigmatizing \\ \hline
complains & expressed concerns about & Evidential \\ \hline
denies & indicates no & Evidential \\ \hline
endorses & agrees & Evidential \\ \hline
noted & mentioned & Evidential \\ \hline
reports & shared that & Evidential \\ \hline
says & describes & Evidential \\ \hline
tells me & confided in me & Evidential \\ \hline
challenging & required additional support & Negative \\ \hline
combative & inviting further consideration & Negative \\ \hline
defensive & navigating significant barriers & Negative \\ \hline
exaggerated & emphasized & Negative \\ \hline
disagreeably & had discomfort with & Negative \\ \hline
deceitfully &  & Negative \\ \hline
deceptively &  & Negative \\ \hline
blatantly &  & Negative \\ \hline
absurdly &  & Negative \\ \hline
alarmingly &  & Negative \\ \hline
insist & is sure of & judgmental \\ \hline
claims & is experiencing & judgmental \\ \hline
evidently &  & judgmental \\ \hline
adamant & is sure of & judgmental \\ \hline
states &  & judgmental \\ \hline
assert & emphasized & judgmental \\ \hline
apparently &  & judgmental \\ \hline
seemingly &  & judgmental \\ \hline
obviously &  & judgmental \\ \hline
\end{tabular}
\caption{Alternative Words/Phrases to Testimonially Injust Terms}
\label{table:alts}
\end{table}

\subsection{Personal Data Survey for Human Experts}
\begin{itemize}
    \item What is your race? (Categories here were adaptations from the US Census) (American Indian or Alaska Native, Asian, Black or African American, Native Hawaiian or Other Pacific Islander, White, Other, Prefer not to say)
    
    \item What is your age range? (Under 18, 18-24 years old, 25-34 years old, 35-44 years old, 45-54 years old, 55-64 years old, 65+ years old)

    \item What is your profession as it relates to the medical care system?
    
    \item What is your primary medical institution you work for? (Categories here were adaptations from: this WGU Blog \footnote{https://shorturl.at/6JQ9B})(Ambulatory Surgical Centers, Birth Center, Blood Banks, Clinic Office, Education Center, Dialysis Center, Hospice, Hospital, Imaging and Radiology Center, Mental Health Center, Addiction Treatment Center, Nursing Home, Orthopedic Center, Urgent Care, Telehealth, Private Practice, Resident, Primary Care, Other)

    \item What is your sex? (Male, Female)
    
    \item What is your geographical region of practice? (Modified from Jagran Josh \footnote{https://www.jagranjosh.com/general-knowledge/regions-of-united-states-complete-list-history-and-importance-1721218579-1})
    (Northeast (Connecticut, Maine, Massachusetts, New Hampshire, Rhode Island, Vermont, New Jersey, New York, Pennsylvania), Midwest (Illinois, Indiana, Michigan, Ohio, Wisconsin, Iowa, Kansas, Minnesota, Missouri, Nebraska, North Dakota, South Dakota), Southeast (Alabama, Kentucky, Mississippi, Tennessee, Florida, Georgia, North Carolina, South Carolina), East Coast/Mid-Atlantic (Maryland, Virginia, Delaware, Washington, D.C., West Virginia), Southwest (Arkansas, Louisiana, Oklahoma, Texas), Westcoast (Alaska, California, Hawaii, Oregon, Washington, Arizona, Colorado, Idaho, Montana, Nevada, New Mexico, Utah, Wyoming))
    
    \item How long have you been practicing? (0-5 years, 6-10 years, 11-15 years, 16-20 years, 21-25 years, 26-30 years, 31+ years, Retired, No longer practicing medicine)

    \item Flip a coin. If you do not have a coin to flip, please use this one provided by Google: \url{https://g.co/kgs/ftbg7pT} ~ {Google Coin Flip}. What did you get? (Heads, Tails)
\end{itemize}

\end{document}